# A survey of image labelling for computer vision applications

Christoph Sager, Christian Janiesch & Patrick Zschech

View supplementary material

Published online: 18 Apr 2021.

Submit your article to this journal

View related articles

View Crossmark data

ARTICLE                                                                                     OPEN ACCESS

# A survey of image labelling for computer vision applications

Christoph Sager [a], Christian Janiesch [b] and Patrick Zschech [c]

[a]Faculty of Business and Economics, Technische Universität Dresden, Dresden, Germany; [b]Faculty of Business Management and Economics, Julius-Maximilians-Universität Würzburg, Würzburg, Germany; [c]School of Business, Economics and Society, Friedrich-Alexander-Universität Erlangen-Nürnberg, Nürnberg, Germany

**ABSTRACT**

Supervised machine learning methods for image analysis require large amounts of labelled training data to solve computer vision problems. The recent rise of deep learning algorithms for recognising image content has led to the emergence of many ad-hoc labelling tools. With this survey, we capture and systematise the commonalities as well as the distinctions between existing image labelling software. We perform a structured literature review to compile the underlying concepts and features of image labelling software such as annotation expressiveness and degree of automation. We structure the manual labelling task by its organisation of work, user interface design options, and user support techniques to derive a systematisation schema for this survey. Applying it to available software and the body of literature, enabled us to uncover several application archetypes and key domains such as image retrieval or instance identification in healthcare or television.

## 1. Introduction

Fuelled by modern sensor technology, broad access to computing power, and the development of user-friendly programming frameworks, the field of computer vision (CV) is currently experiencing a renaissance that results in groundbreaking new applications: cars begin to drive autonomously in real traffic (Grigorescu et al., 2020), medical diagnosis systems support doctors in detecting hard-to-find diseases (McKinney et al., 2020), and intelligent manufacturing plants detect production deficiencies at an early stage (Zschech et al., 2020).

To realise such scenarios, modern CV systems rely on advanced methods from the field of machine learning (ML). Thus, instead of manually defining rules and patterns to execute vision-based tasks, ML models are able to process spatial information in raw image data and learn patterns automatically that are relevant for prediction tasks like recognising, localising, and segmenting visual objects (Janiesch et al., 2021).

The automated learning of patterns in images is usually organised as a supervised learning task. That is, a function is learned that maps an input to an output based on exemplary input-output-pairs (Brynjolfsson & McAfee, 2017). Consequently, robust ML models require large amounts of training data (here: images) as inputs that are tagged with ground truth labels (e.g., object classes) by humans. Hence, creating meaningful labels is a crucial prerequisite for any ML-based CV application as it determines the quality of the model's results. Therefore, the labelling process constitutes a central form of interaction between humans and machines, in which tacit human expert knowledge is projected into the learning base of the CV system.

To date, the labelling process represents a considerable time and monetary effort in the development of ML-based CV systems (Rapson et al., 2018). In particular, the prediction of multiple object classes in varying surroundings makes thousands of high-quality labels for training a necessity to achieve high precision (Szeliski, 2011). For example, the largest available image dataset ImageNet[1] would have needed an estimated 19 years to be labelled by a single person. Schmelzer (2019) states that data pre-processing accounts for more than 80 % of the total project duration and that the market volume for data pre-processing alone is expected to rise from 500 USD million in 2018 to 1.2 USD billion by 2023. To counter these issues and to accelerate this process, interactive visual software is being developed. Its main goals are the reduction of labelling time and the support of users to create high-quality labels. As the labellers will repeat the same action many times a day, even minor improvements can lead to significant time savings.

Until now, no systematic investigation of the state-of-the-art of image labelling software (ILS) has been carried out. In response, we provide the following contributions:

- We systematically capture, aggregate, and describe the prevalent concepts and features of existing reports and implementations of ILS in a systematisation schema to provide guidance for researchers and practitioners.
- We compare seven current open source implementations and describe archetypical applications and domains.[2]

The article is structured as follows. Section 2 introduces the fundamentals of image labelling for supervised ML. In Section 3, we detail the survey process and existing implementations. We provide the synopsis of the field in Section 4 along the lines of annotation expressiveness, degree of automation, organisation of manual work, user interface design, and user support techniques. In Section 5, we present the systematisation as well as guidance for its application. Further, we elaborate on application and domain archetypes, before we close with a summary and outlook in Section 6.

## 2. Fundamentals and related work

As the foundation of our work, we introduce the fundamentals of CV, ML, and image labelling including its process as well as prior surveys on image labelling and ILS.

### 2.1. Computer vision based on machine learning and artificial neural networks

The field of CV is concerned with the acquisition, processing, analysis, and understanding of digital images to generate symbolic or numerical information that can be used, for example, to support automated decision-making. Just as humans use their eyes and brains to understand the world around them, CV attempts to produce the same effect so that computers can perceive and understand an image or a sequence of images and act accordingly in each situation. This understanding can be achieved by disentangling high-level, symbolic information from low-level image features using models built with the help of geometry, statistics, physics, and learning theory (Forsyth & Ponce, 2003; Szeliski, 2011).

Nowadays, CV tasks are increasingly performed by supervised ML algorithms which automatically learn patterns in images by iterating over large training datasets where input-output-relationships are already known (i.e., images with predefined labels of object classes) (Brynjolfsson & McAfee, 2017). Of particular interest are artificial neural networks. Inspired by information processing in biological systems, they consist of multiple interconnected processing units that forward signals using weights and activation functions. Artificial neural networks learn by processing many examples and iteratively adjusting the internal weights according to the difference to the known outcomes (Janiesch et al., 2021).

Advanced processing units are typically organised into multiple layers that can be arranged into deep network architectures. This gives them the capability to process spatial information in raw image data and automatically learn a representation that is relevant for the prediction task, which is commonly known as *deep learning* (LeCun et al., 2015). A widely used network architecture in the field of CV is that of convolutional neural networks. They comprise a series of stages that allow hierarchical feature learning which is a useful principle to condense low-level features into higher-level concepts. For the task of object recognition, this means that the first few layers of the network are responsible for extracting basic features in the form of edges and corners. These are then incrementally aggregated into more complex features in the last few layers resembling the actual objects of interest, such as animals, houses, or cars. Subsequently, these auto-generated features are used for prediction purposes to recognise objects of interest during automated image analysis (Goodfellow et al., 2016).

### 2.2. Need for image labelling and image labelling software

All supervised ML approaches share the need for large amounts of labelled training data. They require labels that are as distinct as possible and avoid ambiguity. In the case of CV, this corresponds to image datasets where the image content is annotated in a machine-readable format.

Thus, the basic idea of image labelling is the construction of a mapping of visual features with semantic and spatial labels to provide a good description of the image content. Existing literature uses *label* and *annotation* interchangeably to describe the outcome of this process. In the following, we speak of ILS, while image annotation software is used synonymously in CV.

The labelling effort is mostly composed of the time spent by human workers and the associated monetary expenditure for wages and infrastructure. A common way to decrease these costs is to support the manual labelling with ILS and take the human more and more out of the process (Fiedler et al., 2019; Said et al., 2017; Son et al., 2018). As Q. Zhang et al. (2015) state, the encoding of the extracted structural knowledge can be another goal. Others such as Russell et al. (2008) "seek to build a large collection of images with ground truth labels to be used for object detection and recognition research." Further, some applications particularly in healthcare create workflows to assist physicians (Son et al., 2018). Summarising, ILS support the users in image labelling and serve several objectives:

(1) labelling a large set of images in a limited amount of time,
(2) supporting the creation of a complete and balanced image dataset,
(3) increasing the labelling efficiency, thereby reducing the human workload,
(4) assuring and maintaining high accuracy while avoiding label noise, and
(5) encoding the extracted knowledge in an efficient and structured way.

### 2.3. Image labelling process

Image labelling is a process that comprises at least five stages: material collection, labelling, postprocessing, quality assessment, and data export.

Each labelling process starts with the *material collection*, which can be differentiated by its type (e.g., image or video) and further available properties such as meta data. If the current material does not suffice or is unevenly distributed, relevant images for model training can be sourced by crowdsourcing, by image scrapping from the Internet, or through the generation of artificial pictures (Zhuo et al., 2019). Subsequently, the actual *labelling* takes place as detailed in Section 4. *Postprocessing* is strongly connected to the task of labelling itself and seeks to improve labelling quality, for example, by label merging (Russell et al., 2008), visual tag refinement (Chowdhury et al., 2018), or rotating and scaling of images for uniform distribution of object locations (Russell et al., 2008). As the performance of learning algorithms depends on the quality of the underlying data, ILS offer ways of *quality assessment* to address class interpretation errors, instance interpretation errors, and similarity errors, which can lead to under- or over-representation creating a bias. Common methods comprise annotating the image multiple times or showing similarly labelled images to the user. The evaluation of image labelling can be performed qualitative by humans or quantitative by metrics (W.-C. Lin et al., 2016). The most common measure is the average accuracy as the share of correct annotations of all relevant objects. Finally, there is the need to store annotations in a structured format to allow for *data export*. As detailed in Section 4.1, there are several competing formats depending on task and domain.

### 2.4. Related surveys on labelling for computer vision applications

When preparing our literature review, we identified six related survey papers that have a slight relation to our topic of interest considering ILS. For better differentiation with our work, we briefly summarise them below.

Hanbury (2008) conducted a survey on approaches for image annotation, while Yan et al. (2008) compare the efficacy of browsing, tagging, and hybrid procedures. However, both reviews have a narrow focus by solely considering the organisation of labelling and the impact of user interface designs. Abdulrazzaq and Noah (2014) provide an overview of content-based image retrieval with a strong focus on image search but also cover medical labelling software. D. Zhang et al. (2012) review automatic image labelling techniques focusing on image retrieval. Dasiopoulou et al. (2011) conducted a survey of semantic image and video annotation software and systematise the labelling process by developing a general input-output model and defining categories for various annotation levels. They have not updated their survey since 2011. Lastly, Gaur et al. (2018) review the state-of-the-art in video annotation.

In summary, none of the surveys provides a current view or was conducted from an application- and domain-independent perspective covering chiefly the task of image labelling or ILS. In particular, there is still no comprehensive comparison of available ILS with their different approaches and methods. Moreover, in most surveys the respective authors analysed their own artefacts as part of the selection raising the question of bias. This is where we add to the field by providing a comprehensive and independent survey on prevalent concepts and features of ILS.

## 3. Survey design

### 3.1. Survey procedure

We followed the guidelines for systematic literature analysis established by vom Brocke et al. (2015), but we also considered advice by Kitchenham and Charters (2007) and Webster and Watson (2002). First, we limited the scope of the survey, and then we systematised the topic using identified annotation concepts. We carried out the literature search using selected databases (see Table 1), supplemented by a forward and backward search. We used inclusion and exclusion criteria according to Kitchenham and Charters (2007) to restrict the results to a manageable size.

We summarised the results of the literature search by qualitative and quantitative means to identify concepts and features of image labelling. We mapped the results to the identified applications and domains for image labelling, showing which concepts and features are suitable in which context. In addition, we compared a selection of ILS in a concept matrix as proposed by Webster and Watson (2002).

Table 1. Application of the search strategy with inclusion & exclusion criteria.

| Database | Hits | Not German or English | Not covering labelling software or features | Already included in literature corpus | Different application context | Only specific ML algorithm | Included in literature corpus |
| --- | --- | --- | --- | --- | --- | --- | --- |
| ACM Digital Library | 115 | 0 | 49 | 1 | 31 | 5 | 29 |
| AISeL | 110 | 0 | 85 | 13 | 3 | 6 | 3 |
| arXiv | 177 | 0 | 110 | 2 | 17 | 31 | 17 |
| ASC | 86 | 1 | 47 | 1 | 10 | 13 | 14 |
| IEEE Xplore | 101 | 4 | 31 | 1 | 24 | 8 | 33 |
| PubMed | 67 | 0 | 42 | 3 | 5 | 6 | 11 |
| ScienceDirect | 76 | 0 | 60 | 0 | 9 | 3 | 4 |
| Springer Link | 78 | 0 | 42 | 4 | 15 | 8 | 9 |
| Backward Search | 4 | . | . | . | . | . | 4 |
| Forward Search | 3 | . | . | . | . | . | 3 |
| Total | 817 | −5 | −466 | −25 | −114 | −80 | 127 |

### 3.2. Search scope and search strategy

The scope of our search is *single purpose* ILS that help users to annotate images for the training of CV algorithms. We performed our literature review sequentially. That is, we concluded the literature search before the analysis (Kitchenham & Charters, 2007). We aim for comprehensive coverage and selected the databases ACM Digital Library, AISeL, arXiv, ASC (EBSCOhost), IEEE Xplore, ScienceDirect, and SpringerLink as they cover the prime outlets for ILS and ML applications. As CV is also an emerging technology in healthcare (Giger, 2018), we included the PubMed database as well. We searched title, keyword attributes, and abstract without restrictions on publication year, impact factor, or citation count.

Our final search term was "*(label\* tool) OR (annotat\* tool) OR (image label\*) OR (image annotat\*)*". When databases did not support wildcards and truncation, "labeling" and "annotation" were used in their full spelling. For most databases, we had to restrict the search further to keep the results to a manageable amount as they contained a significant number of irrelevant publications in the fields of education, marketing, and healthcare. We included papers in German or English language where the paper covers ILS or their concepts and features. We excluded duplicates and prior versions of updated derivatives if the paper covers mainly a different application context and does not provide additional concepts for image labelling (e.g., natural language processing), or if the paper covers mainly a specific ML algorithm or method. See Appendix A.1 for the detailed search strings and their results.

We carried out the final search in January 2020. Table 1 shows the hits as well as reasons for the exclusion of the papers for each database. A further backward and forward search resulted in 7 additional papers. In total, 127 publications remained for in-depth analysis. See Appendix A.2 for the result set.

### 3.3. Survey statistics

All 127 identified publications were published from 1999. Figure 1 conveys that research has increased significantly over the past five years. We suspect that the rising trend of ML applications for CV resulted in research for software supporting the annotation of large image datasets.

Much labelling software was developed for specific applications or at least to be used in selected domains only. The most frequent domain is healthcare (24 publications), which may be due to the inclusion of the PubMed database. Here, the automated analysis of images is a trending topic in the area of computer-aided diagnosis (CADx). Additional application areas are the analysis of television data (7) and the annotation of street scenes for autonomous driving (7). Also, the evaluation of surveillance videos or the mapping of aerial images is covered in multiple publications.

During the analysis, we assigned keywords to the corpus. For the distribution see Figure 2. Most of the publications cover only image labelling (58). However, some also support the labelling of video data (24), while others only focus on a specific aspect or method related to labelling (38). There is a large variability of features in image labelling. While 26 publications support semi-automated labelling, more and more features concerned with semantics (16) or user interaction (10) were developed in recent years. Also, the Internet and social media influenced labelling software by enabling collaboration and crowdsourcing or crawling (relational) meta data from online sources. Today, the implementation of privacy-preserving methods or modules for the explanation of machine annotations is still a minor field. We expect more research on these topics in the future.

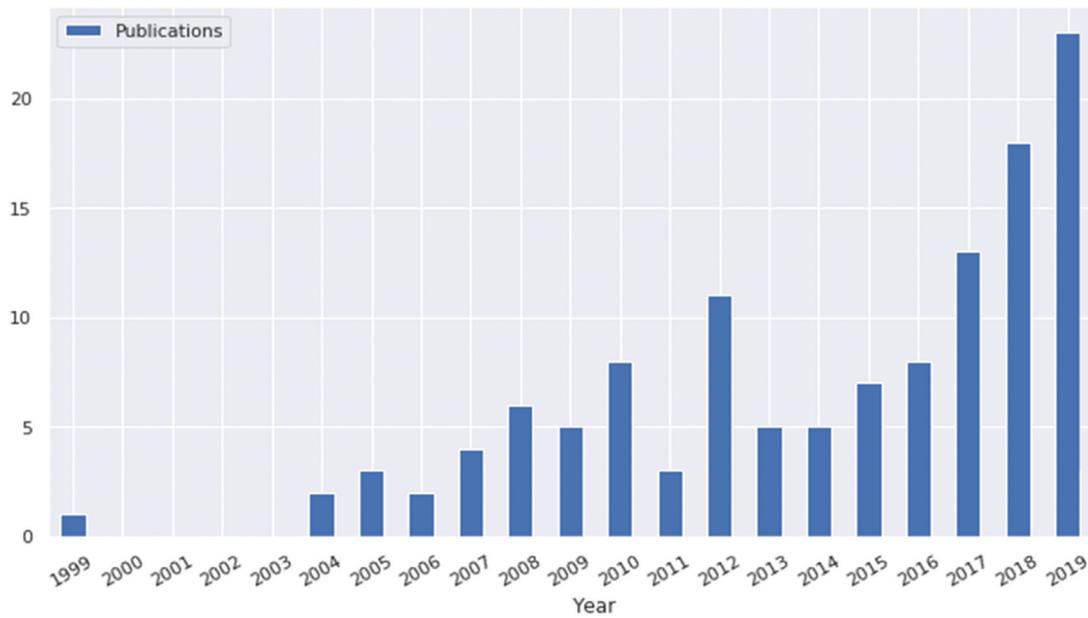

Figure 1. Temporal distribution of publications from 1999 to 2019.

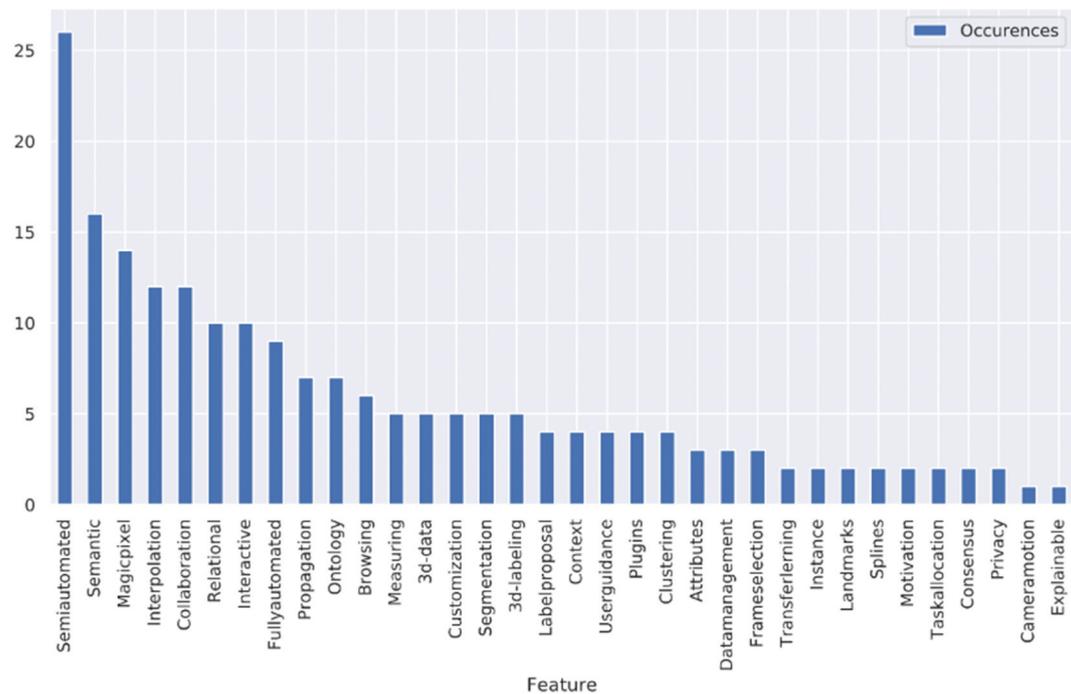

Figure 2. Distribution of features in ILS.

### 3.4. Identified image labelling software

In our literature search, we identified 58 ILS that support image labelling and 24 ILS that support video annotation. While many ILS have been described in literature, only a small subset is still available for use. Therefore, we created a subset of literature based on two inclusion criteria:

(1) The software is still publicly available and can be used.
(2) The software has a certain degree of universality; that is, it has not been developed for a unique application context.

After applying the criteria, only six ILS remained for a detailed analysis. We excluded most software due to the websites being defunct or for them being single-purpose tools for medical diagnosis. In addition, we included the ILS labelImg and Microsoft's Visual Object Tagging Tool (VoTT), which have no academic precursors but are very popular in the research community. Table 2 provides an overview of the selected software. We provide more details on the concepts and features used in these ILS in Section 5.1.

Table 2. General purpose image labelling software.

| Year | Name | GitHub URL | Reference |
| --- | --- | --- | --- |
| 2008 | LabelMe | .../CSAILVision/LabelMeAnnotationTool | Russell et al. (2008) |
| 2015 | labelImg | .../tzutalin/labelImg | - |
| 2016 | VIA | .../vgg/via/ | Dutta and Zisserman (2019) |
| 2017 | CVAT | .../opencv/cvat/ | Said et al. (2017) |
| 2017 | VoTT | .../microsoft/VoTT/ | - |
| 2018 | Web Annotation | .../mpizenberg/annotation-app/ | Pizenberg et al. (2018) |
| 2019 | ImageTagger | .../bit-bots/imagetagger/ | Fiedler et al. (2019) |

LabelMe is one of the oldest and most common ILS in this field of research. It was created due to the lack of available online datasets with annotated images (Russell et al., 2008). The developers decided on a collaborative approach by publishing images and asking Internet users to annotate them without limiting shapes or words. labelImg allows for quick ad-hoc annotation of bounding boxes and the handling of multiple object classes. The Visual Geometry Group Image Annotator (VIA) was developed as a plain website. It supports the labelling of images and videos, as well as voice records (Dutta & Zisserman, 2019). Intel's CV Annotation Tool (CVAT) is a browser-based open source application for both individuals and teams that supports different work scenarios. Its main function is to provide users with convenient annotation instruments. VoTT provides end-to-end support for generating datasets and validating object detection models from video and image assets. Web Annotation's focus is on constituting training datasets for ML algorithms with an end-user focus. It is embeddable in Amazon Mechanical Turk (Pizenberg et al., 2018). Lastly, Fiedler et al. (2019) developed ImageTagger with a focus on collaborative work and adaptability to specific use cases. The primary use case of the software is training robots to detect footballs, but it can be customised for other scenarios.

## 4. Concepts and features of image labelling software

In our analysis of the current state-of-the-art of ILS, we the examined annotation expressiveness of the labels, the degree of labelling automation, the organisation of manual labelling, user interface designs as well as labelling support techniques.

### 4.1. Annotation expressiveness

Image labels vary strongly along the two axes of semantic and spatial granularity. These levels of detail correspond with the expressiveness of the resulting annotations.

#### 4.1.1. Semantic expressiveness

The type of annotation vocabulary has a substantial influence on the design of ILS. We can distinguish several types of annotation: free text, keywords, taxonomies, ontologies, and instance labels, which all share the same objective of describing images by explicating their semantic meaning.

Free text annotations have no pre-defined structure and allow humans to write full sentences to describe the image. They are often used for scene descriptions or automated annotations with meta data. The other types typically rely on a defined vocabulary. Keywords, also called tags, label resources in a subjective and often associative way using a single term (Toyama & Konishi, 2008). While some tools restrict images to a single label creating mutually exclusive categories (classification), most ILS allow for associating a picture with multiple tags for multi-label classification. Taxonomies structure terms in a hierarchical tree and allow the inference of further labels, while ontologies provide a specification of a conceptualisation (Kim et al., 2013). Therein entities (concepts) are connected by relationships complying with some global rules such as equivalence or negation (Hanbury, 2008). For example, WordNet is a collection of 117k synonym sets (synsets) that is used by various ILS (Hanbury, 2008; Z. Lin et al., 2011; Yao et al., 2007). Lastly, labelling individual instances of a class can add further meaning to an image description. This can be useful for surveillance to identify individuals (Lee et al., 2019). See Figure 3 for an overview of the semantic concepts.

Some ILS also allow for further characterising entities by adding attribute values to objects (Bianco et al., 2015). In contrast to closed vocabularies, Web-based collaborative tagging systems allow users to define and add labels without limitation, resulting in so-called folksonomies. They lead to very detailed, yet at times ambiguous descriptions (Russell et al., 2008).

The storage of annotation semantics is difficult if the application goes beyond a few unrelated classes. Examples of widespread formats are the RDF- and OWL-format, as well as the MPGEG-7 file (Athanasiadis et al., 2007; Giro-i-Nieto & Martos, 2012; Theodosiou et al., 2009). Many ILS use proprietary formats and they are often defined for a single domain. For example, the Annotation and Image Markup (AIM) is specific for medical labelling (Rubin et al., 2009), and it is compatible with OWL and the healthcare-specific Digital Imaging and Communications in Medicine (DICOM) standard.

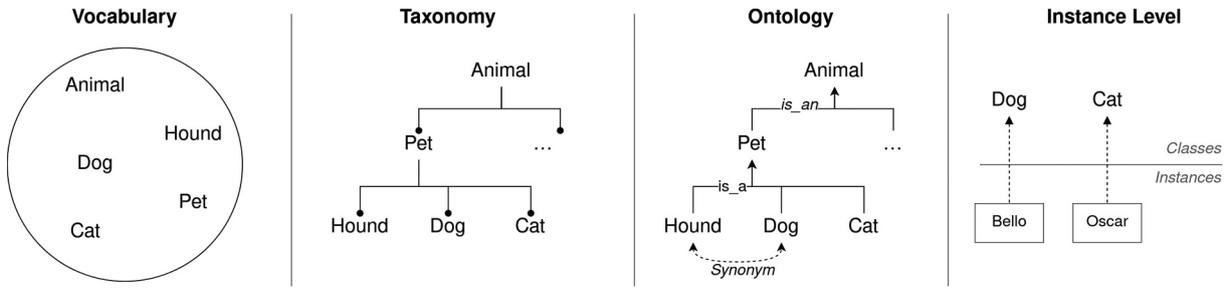

Figure 3. Illustration of various semantic concepts.

### 4.1.2. Spatial expressiveness

Annotation granularity is another important distinction (Dasiopoulou et al., 2011). Labels can refer to the whole image (scene annotation) or specific spatial segments (region/segment-based annotation). Giro-i-Nieto and Martos (2012) define this as (i) global labelling for single-label or multi-label image classification and (ii) local labelling for object detection.

Here, a major challenge is to differentiate if two tags refer to the same or two different objects in the image scene (Li & Yeh, 2018). Therefore, local labels describe multiple identified objects in the same image. They can be further differentiated by their form. Most ILS support labelling with so-called bounding boxes that can come in other geometric shapes such as circles and ellipses. Moreover, there exist further options such as polygons, segments/masks, points, and (poly-)lines (see Figure 4).

In face detection applications, for example, eyes are often marked with key points. Lines or polylines are common to support edge detection, for example, in autonomous driving (Hu et al., 2008). Polygons can express more detailed object shapes, and spatial labelling creates segments for each object. These masks are created by redrawing the area covered by the object and therefore obtaining pixel-precise representations, including holes not belonging to the object.

While spatial information adds more expressiveness to the annotation, the labelling task also becomes sequentially more complicated when moving from scene classification over object detection to object segmentation (Rapson et al., 2018). Most ILS save pixel coordinates in some form of XML structure. Widespread formats are Pascal VOC or structured JSON-files as used by the COCO dataset.

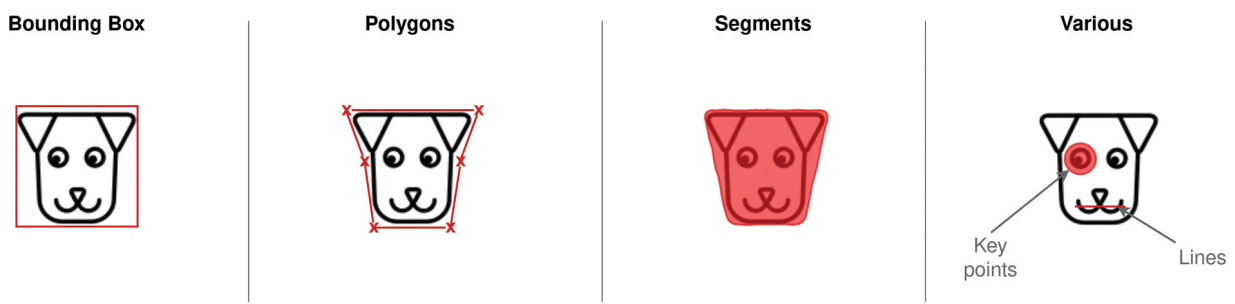

Figure 4. Visualisation of the different spatial labelling types.

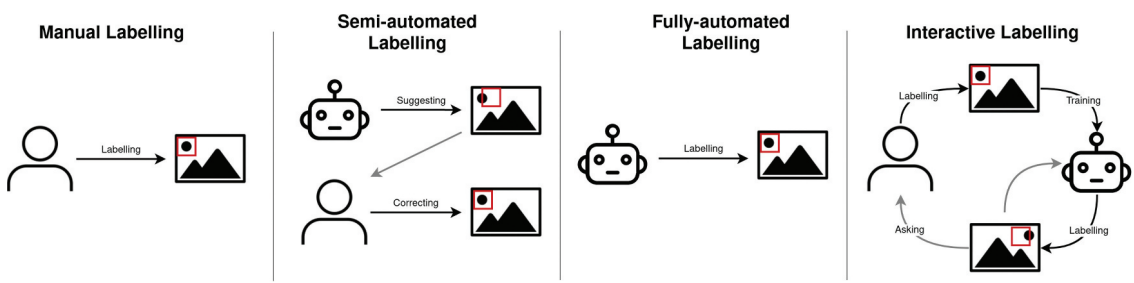

Figure 5. Automation degrees of the labelling process.

### 4.2. Degree of automation

The involvement of human effort in the manual labelling process and its reduction has been the primary topic of research in recent years. Semi-automated labelling limits labour to correcting proposed annotations, while in fully automated labelling, there is no human involvement for labelling at all. Interactive labelling has emerged as an integration of the former three. See Figure 5 for an overview.

#### 4.2.1. Manual labelling

The main objective of manual ILS is to provide a simple and efficient user interface that is adapted to the cognitive ability of humans so that the user can label fast and accurately. Newer ILS also focus on features such as collaboration or specific application contexts. Some ILS avoid distracting the user by limiting the number of visible concepts, while others place related instruments next to each other, so that the user does not have to switch sights (Tsai et al., 2015). Annotation time can be further decreased by labelling support techniques (see also Section 4.5).

#### 4.2.2. Semi-automated labelling

To ease the burden of labelling large amounts of images by hand, it is possible to automate certain parts of the process: First, a CV algorithm performs a preliminary annotation. Second, the human user reviews and corrects the proposed labels. Consequently, the role of the human changes to supervision with the sole tasks of filtering, selecting, and updating.

Larumbe-Bergera et al. (2019) and Bianco et al. (2015) significantly reduce the labelling time with semi-automated labelling in different domains. Further, Russell et al. (2008) discuss the possibility to enhance their database by integrating semi-automated labelling and image searches to fill up underrepresented classes. Some ILS incorporate a different visualisation for automatically created annotations by using different colours or dashed borders and provide a locking mechanism to prevent further change of manual annotations by the automatic algorithm (Philbrick et al., 2019). Kim et al. (2013) see semi-automation as a good trade-off between human effort on the one hand and accuracy as well as broad scene understanding, on the other hand. Beugher et al. (2018) even argue that the only way to ensure high accuracy in all circumstances is to integrate a minimal amount of human intervention.

#### 4.2.3. Automated labelling

Fully automated labelling describes ILS that associate images with labels without human intervention based on an existing model of classes. Most of the algorithms used for this purpose are based on some form of clustering. In addition, transfer learning can be used to support the generation of the necessary classification models (Athanasiadis et al., 2007; Iakovidis et al., 2014). Uricchio et al. (2017) build a semantic space of textual and visual features, where similar images appear close to each other. The algorithm extracts the features of new images and compares them to the classification of images with similar features. Aljundi et al. (2017) use clustering to detect actors in movies and TV series. They rely only on a small initial portrait dataset extracted from a movie database, which they transfer to video scenes without human intervention. Zhuo et al. (2019) use annotated maps, leveraged with remote sensing data for fully automatic image labelling. They conclude that fully automated labelling can sometimes even yield better results than manual image labels, as the latter is always subjective to human bias and human error.

#### 4.2.4. Interactive labelling

Interactive labelling, sometimes also called active learning or "annotation by iterative deep learning" (Philbrick et al., 2019, p. 577), is part of the human-in-the-loop methodology. Its objective is to reduce the limits of fully automated labelling through targeted user interaction (Nadj et al., 2020). The method lets the user label an initial batch of examples, trains a model, and then continuously asks the user for corrections. L. Zhang et al. (2008) developed an active user feedback strategy, which minimises the errors in subsequent labelling iterations and maximises the expected information gain. The interactive annotation and segmentation tool (iSeg) is an example of this kind of interaction where user intervention is limited to either relocation, shape adjustment, or the creation of a completely new annotation. IGAnn (Interactive imaGe ANNotation) supports interactive labelling combined with a hierarchical approach (Chiang, 2013). The user associates a single label with all relevant images displayed in a grid, a classifier is trained in the background and returns a list of images with the highest confidence for the next iteration.

### 4.3. Organisation of manual labelling

The human task of image labelling can be performed in several ways. In the following, we detail single user labelling, collaborative labelling, and crowdsourcing. Further, Hanbury (2008) distinguishes annotation parties as an effective means if done by people together in a brief period of time and possibly close location.

#### 4.3.1. Single user

Single user labelling is the most basic form of image labelling and denotes a single user labelling images by him- or herself. The result is very likely to contain a bias especially for borderline cases as the reliability of labels is not immediately clear. This mode is supported by most tools and can be conducted offline as no synchronisation is necessary.

### 4.3.2. Collaboration

Collaborative labelling enabled sharing annotation decisions over multiple people, and therefore, is less prone to error and bias. In areas with high inter-expert variability (e.g., healthcare), a collaborative approach can increase the consensus (Mata et al., 2017). If multiple persons label the same images, the tasks can be distributed either parallel or sequential. Using the parallel strategy, one receives two independent answers, which then must be merged. Using the sequential strategy, the second user will see the label of his or her predecessor and can decide if it should be changed. This approach is susceptible to bias from the initial annotation.

### 4.3.3. Crowdsourcing

Crowdsourcing via public platforms that distribute paid microtasks is another option that enables extensive scalability. Hughes et al. (2018) analysed a large platform-substituting medical experts and found that there exists some kind of "wisdom of the crowd" by using highly redundant annotations and merging them with various consensus strategies. These tasks comply with Fitt's law of speed-accuracy trade-offs in human pointing tasks: Longer click periods lead to higher accuracy. While having comparable accuracy to individual experts, crowdsourcing is still slower and less accurate than a consensus of trained personal. It is only useful where massive scalability is of advantage. Besides, annotation complexity and worker fluctuation must be considered for high-quality labels.

### 4.4. User interface design

The user interface design of ILS can be split into two annotation paradigms: tagging and browsing. For tagging, the ILS presents the user a single picture and asks to enter all relevant tags that he or she associates with that image. For browsing, the ILS presents a single term and a whole set of images, often arranged in a grid, and asks the user to select all images, which he or she can connect to the keyword.

While tagging single images is by far the more common approach, there are several ILS with a browsing interface (Chiang, 2013; Münzer et al., 2019; Volkmer et al., 2005), and some ILS that support browsing as an additional user view (Dutta & Zisserman, 2019; Giro-i-Nieto & Martos, 2012). The grid view is useful for quickly reviewing many images, given that the objects of interest are not too small. However, browsing is not very flexible and requires the user to define a closed vocabulary beforehand. Often, it only supports image-level classifications, since drawing spatial annotations on small thumbnails is difficult (see also Figure 6). In contrast, the annotation time is usually shorter, because the user only focuses on a single concept. Yan et al. (2008) see both techniques as complementary. They conclude that tagging is more suitable for infrequent keywords (e.g., individual instances) while browsing works best for common keywords (e.g., person or car). Further, they propose a mixed approach of intermitting browsing and tagging.

Group-based labelling is another method of iterative single-concept labelling using a browsing interface. It clusters the images beforehand by visual features and then asks the user to associate a whole group of similar images with a single label. The method does not require any a priori knowledge about the data, but inherently introduces noise as not all images of the group belong to the chosen concept. Wigness et al. (2018) develop a hierarchical clustering method, which tries to represent the coarse-to-grain structure of visual concepts. Hence, labels close to the root of the hierarchy represent broad concepts (e.g., "animal"), and leaves contain very specific labels (e.g., breed "German shepherd"). User interaction is restricted to specifying a label or telling the algorithm that the group is too dissimilar. Their classification accuracy is up to 15 % higher after 15 minutes compared to traditional approaches.

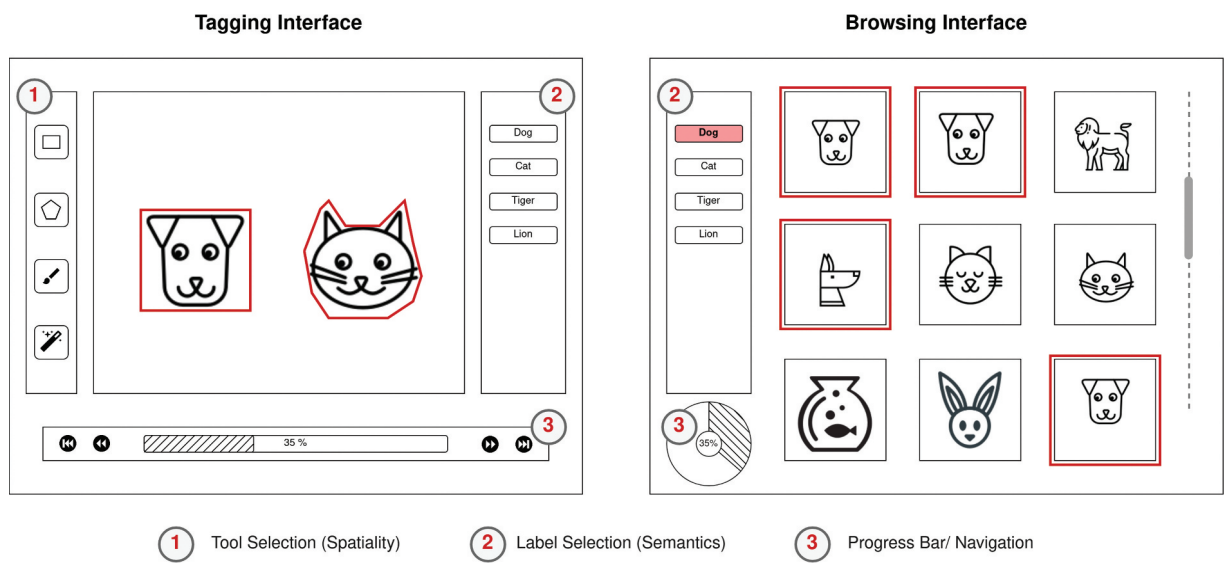

Figure 6. User interface comparison of the two design paradigms.

In general, there is an agreement on a straightforward design and setup, for example, as Web applications (Dutta & Zisserman, 2019; Fiedler et al., 2019; Russell et al., 2008; Volkmer et al., 2005). For example, Dutta and Zisserman (2019) refrain from adding further features at the expense of an increased complexity. In contrast, Pizenberg et al. (2018) allow extensive configuration so that the users can adapt the interface to their individual needs (e.g., add instruments or change brush colour for different concepts). This is an essential point as a lot of ad-hoc ILS were created because existing ones did not fit the purpose (see Fiedler et al., 2019).

### 4.5. Image labelling support techniques

As manual image labelling is time-consuming and costly, many techniques have been developed to support the human task, namely, label proposal, magic pixel, intelligent scissor, mask dragging, and label propagation.

#### 4.5.1. Label proposal
Suggesting likely relevant labels renders simple assistance to the user but may introduce a bias. Proposals can be ordered by class frequencies of the dataset (Toyama & Konishi, 2008). More complex strategies also incorporate already given labels of the image or semantic relations defined in an ontology (Adebayo et al., 2016).

#### 4.5.2. Magic pixel
The spatial annotation of polygons and segments is very time-consuming and exhausting for human labellers. Therefore, many approaches exist to support or automate the creation of object boundaries. Due to a lack of automatic algorithms, most authors use interactive image segmentation. The idea is that the user selects the object with minimal effort by only giving some rough information about the region of interest. Breve (2019) identifies three overall categories of user interaction: (i) the user loosely traces desired boundaries, (ii) the user marks parts of the object of interest and/or background, and (iii) the user loosely places a bounding box around the object of interest.

The most common method is known as the magic wand function. It corresponds to Breve's second category, and the user simply selects a single pixel (superpixel) inside the selected object. This superpixel groups perceptually similar pixels to create visually meaningful entities to reduce the cost for subsequent processing (Stutz et al., 2018). It can be configured by specifying the colour range and tolerance but has trouble detecting objects with multiple colours (Wu & Yang, 2006). An example of Breve's first category of boundary tracing is Ratsnake (Iakovidis et al., 2014), where the user draws a coarse polygon, which is refined by an active contour model. The snake model can be optimised by including prior knowledge of the class shapes. Giro-i-Nieto and Martos (2012) present a partition tree-based technique, which merges the most similar neighbouring regions. They offer various initial selection strategies: bounding box (Breve's category three), single-click, mouse-overs, and scribbles (category two). Their ILS also allows the user to navigate inside the partition tree (shrinking/enlarging the segment) by rotating the mouse wheel. See Figure 7 for an overview of various magic pixel features.

#### 4.5.3. Intelligent scissor
The intelligent scissor is an instrument similar to superpixels. It automatically suggests contours that fit high-contrast boundaries and calculates the shortest path in a graph weighted by gradient magnitude. The user then only provides a coarse line that will snap to the nearest boundary. Live wire tools show where the contour would be, thus giving the user more control over the labelling task (Little et al., 2012).

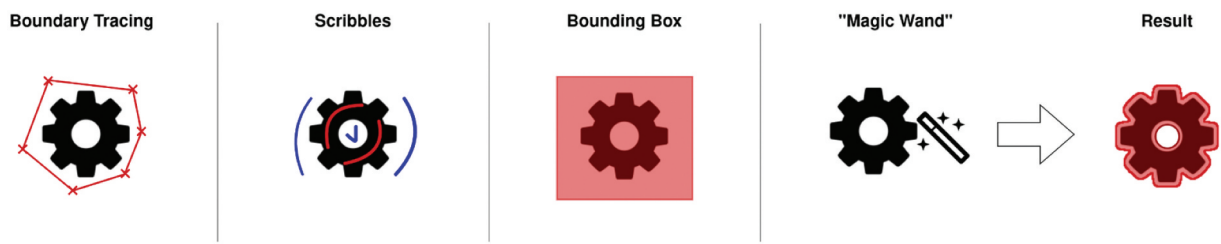

Figure 7. Examples of various magic pixel features.

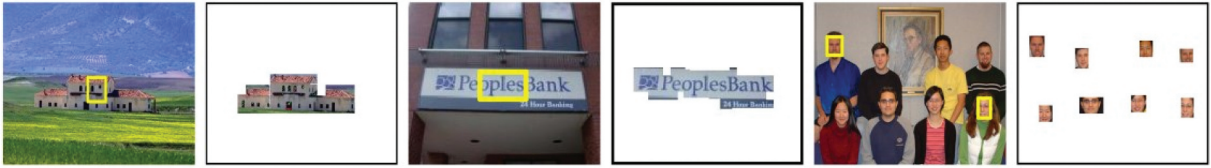

**Figure 8.** Example annotation with SmartLabel of Wu and Yang (2006).

### 4.5.4. Label propagation

The term label propagation is not consistently used, and implementation varies. Most propagation methods are based on some form of clustering. Ivanov et al. (2010) use hierarchical *k*-means clustering of visual features to detect similar pictures. When the user annotates an object on a new image, the system automatically propagates this annotation to similar objects from the database using a duplicate detector. Verma and Jawahar (2017) follow a similar concept but additionally incorporate image-to-label similarities. Based on the assumption that similar images share similar labels, they also consider semantic neighbours while propagating labels. With SmartLabel (Wu & Yang, 2006), the user only has to label a sub-region of an object, and an algorithm uses this knowledge to grow the region and label the remaining objects in the same image (see Figure 8 for examples).

We provide a discussion of support techniques for video labelling in Appendix A.3.

### 4.6. Gamification

Humans tend not to enjoy data labelling for longer periods because the task is a monotonous activity. While commercial projects can compensate this with monetary incentives, voluntary crowdsourcing projects face a major challenge to acquire labellers. A popular solution to this problem and also to diversify or improve label quality is the integration of gamification aspects that motivate with extrinsic rewards.

The ESP game (von Ahn & Dabbish, 2004) was one of the earliest examples. At the time, it aimed at providing semantic descriptions "for most images of the Web." The ILS generates useful labels in an entertaining way by matching two players, which have to find the same terms independently to describe a given picture. In five years, the game collected 10 million image captions. Ho et al. (2010) enhance this approach by adding a competitive player to their game KissKissBan. This player, "the blocker", sets the taboo words with the goal of preventing the couples from reaching consensus. This constellation naturally impedes cheating (i.e., using words not related to the image) and brings up more diverse sets of image labels. ClueMeIn (Harris, 2018) builds on the ESP game but changes the gameplay to obtain more detailed image labels. It presents both players with multiple similar images and asks them to come up with terms that distinguish them from another.

Another method is only to use gamification principles in otherwise common ILS. For example, BUBL augments selecting the regions of interest by overlaying the picture with a superimposed bubble wrap interface since "popping bubbles seems to be an enjoyable and fun activity" (Galleguillos et al., 2009, p. 3). By iteratively varying the bubble/hexagon size, even fine-grained object boundaries can be collected using this method.

## 5. Discussion of survey

In the following, we discuss our findings from the survey and provide an overview as well as guidance of concept and feature coverage by the current ILS that we identified in Section 3. Further, we discuss its application in encountered application archetypes and the predominant business and research domains.

### 5.1. Systematisation and guidance for image labelling software selection

In the following, we map the identified image labelling features and concepts to a schema to structure a survey of current ILS. To keep the systematisation concise and clear, we did not include some marginal concepts found in specific ILS only. Table 3 summarises the schema.

#### 5.1.1. Comparison of image labelling software

We conducted the comparison of the identified ILS according to the nine dimensions identified in Table 3. For each dimension, we checked all characteristics as well as their occurrences and noted the results in a binary fashion. The result of the comparison is available in Table 4.

It is evident from our discussion that every application poses different requirements on ILS. Hence, this systematisation and comparison does not intend to identify a best in class but to provide a balanced decision template for ILS selection based on given requirements. It can be applied to new market entrants as well. To give further guidance for selecting the most suitable ILS in a given situation, we provide a straightforward decision tree (see Figure 9). It comprises the latter seven dimensions as defined in Table 3 and offers the reader a simple step-by-step process to identify concepts and features relevant to his or her particular application context.

Table 3. Schema for image labelling software systematisation.

| Dimension | Characteristic | Explanation |
| --- | --- | --- |
| General | Active, open source | *Active* denotes projects that have not been abandoned. *Open source* signifies that the source code is available. |
| Material | Image, video | Input material can be either *image* or *video* data. |
| Semantic expressiveness | Openness of vocabulary, structure, instance labelling | *Openness of vocabulary* means that the vocabulary is extensible. Structure comprises different types of vocabulary organisation such as *keyword, ontology, taxonomy*, and *attributes* as outlined above. *Instance labelling* signifies said support. |
| Spatial expressiveness | Scene annotation, object detection | *Scene annotation* signifies the ability to classify images. *Object detection* comprises the techniques *bounding box, circle, ellipse, polygon, key point, line, polyline*, and *segments* as outlined above. |
| Degree of automation | Manual, semi-automated, fully automated, interactive | Automation distinguishes the ability for the user to label in a *manual, semi-automated, fully automated*, or *interactive* fashion as outlined above. |
| Organization of labelling | Single user, collaboration, crowdsourcing | *Single user* means that the ILS does not provide support for *collaboration* in labelling with other users or *crowdsourcing* the image and label to the community. |
| User interface | tagging, browsing | The user interface allows for labelling individual images (*tagging*) and/or multiple images (*browsing*) at a time. |
| User support techniques | image labelling, video labelling | *Image labelling* comprises the techniques *label proposal, magic pixel* (for example, based on bounding box, boundary tracing, scribbles, or magic wand), *intelligent scissor*, and *label propagation* as outlined above. For *video labelling* techniques see Appendix A.3. |
| Gamification | | Signifies that the software supports some form of *gamification* for user engagement. |

### 5.2. Image labelling application archetypes

While we focus on general-purpose ILS, we found that there are at least five distinct archetypes of image labelling application with distinct annotation and process requirements.

#### 5.2.1. Complex scene understanding

To understand complex image scenes, it is vital to extract fine-grained information to obtain correct semantics as well as relational information between the objects. Consequently, ontologies (such as WordNet) should be used, and instance recognition can be necessary, for example, to follow the same object over a sequence. For the spatial labelling, bounding boxes might not be sufficient as shapes often diverge from simple rectangles, leading to polygons or segmentation as a better option. Due to the high level of spatial detail, a tagging interface is recommended for fine-grained labelling. The use of crowdsourcing can lead to a more diverse set of objects and increases the model's robustness. Still, occasional manual trimming of the annotation vocabulary is advisable. Semi-automated labelling, as well as magic pixel and label interpolation features, can be used to speed up the labelling process.

#### 5.2.2. Diagnosis and classification

These tasks typically comprise only a few different classes and focus mostly on the identification of correct labels. Therefore, a simple vocabulary of defined terms is sufficient, and spatial annotations (e.g., segmentation) are recommended to gain additional information about the objects (position and shape). For subjective tasks (e.g., the interpretation of skin lesions), collaboration and redundant labelling are recommended to share decision making. ILS should further provide modules for user and rights management as well as a review mode that allows experts to review (automatic) labels. In addition, automatic consensus-finding and semi-automatic labelling can support the user and improve the accuracy. For decisions with significant consequences, the user should stay in the loop and supervise model outcomes to cover for edge cases.

#### 5.2.3. Image retrieval

Usually, databases should be structured by semantic concepts (such as scene descriptions). Therefore, it makes sense to use ontologies for large annotation vocabularies, which can be used for label suggestion (to keep consistency) and visual tag refinement in postprocessing. Spatial annotation, however, is usually not necessary, as label position and shape often are not crucial for simple image search. The ILS interface should offer a browsing view to foster the quick labelling of common concepts. If semi-automated labelling is used, browsing can also help to review the suggested labels.

#### 5.2.4. Instance identification

The identification of instances is an unusual and infrequent use case. Consequently, there exist some ILS solely for this purpose, for example, for surveillance. Still, features of general ILS can be identified to support this application. Face identification, for instance, tries to recognise a specific person from a set of metrics based on landmarks and polylines. This can be supported by pre-trained models using semi-automated labelling and magic pixel features. The tagging view should be used as it allows a detailed image analysis and supports the labelling of multiple different classes and instances. In the case of CCTV camera data, label interpolation and object tracking can help to reduce the manual labelling effort (see also Appendix A.3).

Table 4. Comparison of selected image labelling software.

| | | LabelMe | labelImg | VIA | CVAT | VoTT | Web Annotation | Image Tagger |
|---|---|---|---|---|---|---|---|---|
| General | Active | ✓ | ✓ | ✓ | ✓ | ✓ | ✗ | ✓ |
| | Open Source | ✓ | ✓ | ✓ | ✓ | ✓ | ✓ | ✓ |
| Material | Image | ✓ | ✓ | ✓ | ✓ | ✓ | ✓ | ✓ |
| | Video | ✓ | ✗ | ✓ | ✓ | ✓ | ✗ | ✗ |
| Semantic Expressiveness | Open Vocabulary | ✓ | ✓ | ✓ | ✗ | ✓ | ✗ | ✓ |
| | Structure | Taxonomy, attributes | Keyword | Keyword, attributes | Keyword, attributes | Keyword | Keyword | Keyword |
| | Instance Labelling | ✗ | ✗ | ✓ | ✓ | ✗ | ✗ | ✗ |
| Spatial Expressiveness | Scene Annotation | ✗ | ✗ | ✓ | ✓ | ✗ | ✗ | ✗ |
| | Object Detection | Bounding box, polygon | Bounding box | Bounding box, circle, ellipse, polygon, key point, polyline | Bounding box, polygon, point, polyline | Bounding box, polygon | Bounding box, polygon, line | Bounding box, polygon, line |
| Degree of Automation | Manual | ✓ | ✓ | ✓ | ✓ | ✓ | ✓ | ✓ |
| | Semi-automated | ✗ | ✗ | ✗ | ✓ | ✓ | ✗ | ✗ |
| | Fully Automated | ✗ | ✗ | ✗ | ✓ | ✗ | ✗ | ✗ |
| | Interactive | ✗ | ✗ | ✗ | ✗ | ✗ | ✗ | ✗ |
| Organization of Labelling | Single User | ✓ | ✓ | ✓ | ✓ | ✓ | ✓ | ✓ |
| | Collaboration | ✓ | ✗ | ✗ | ✓ | ✗ | ✓ | ✓ |
| | Crowdsourcing | ✗ | ✗ | ✗ | ✗ | ✗ | ✗ | ✗ |
| User Interface | Tagging | ✓ | ✓ | ✓ | ✓ | ✓ | ✓ | ✓ |
| | Browsing | ✗ | ✗ | ✓ | ✗ | ✗ | ✗ | ✗ |
| User Support Techniques | Image Labelling | Label proposal, magic pixel | ✗ | ✗ | Magic pixel | ✗ | ✗ | ✗ |
| | Video Labelling[3] | Label interpolation, camera motion estimation | ✗ | ✗ | Automatic frame selection, label interpolation | Automatic frame selection | ✗ | ✗ |
| Gamification | | ✗ | ✗ | ✗ | ✗ | ✗ | ✗ | ✗ |

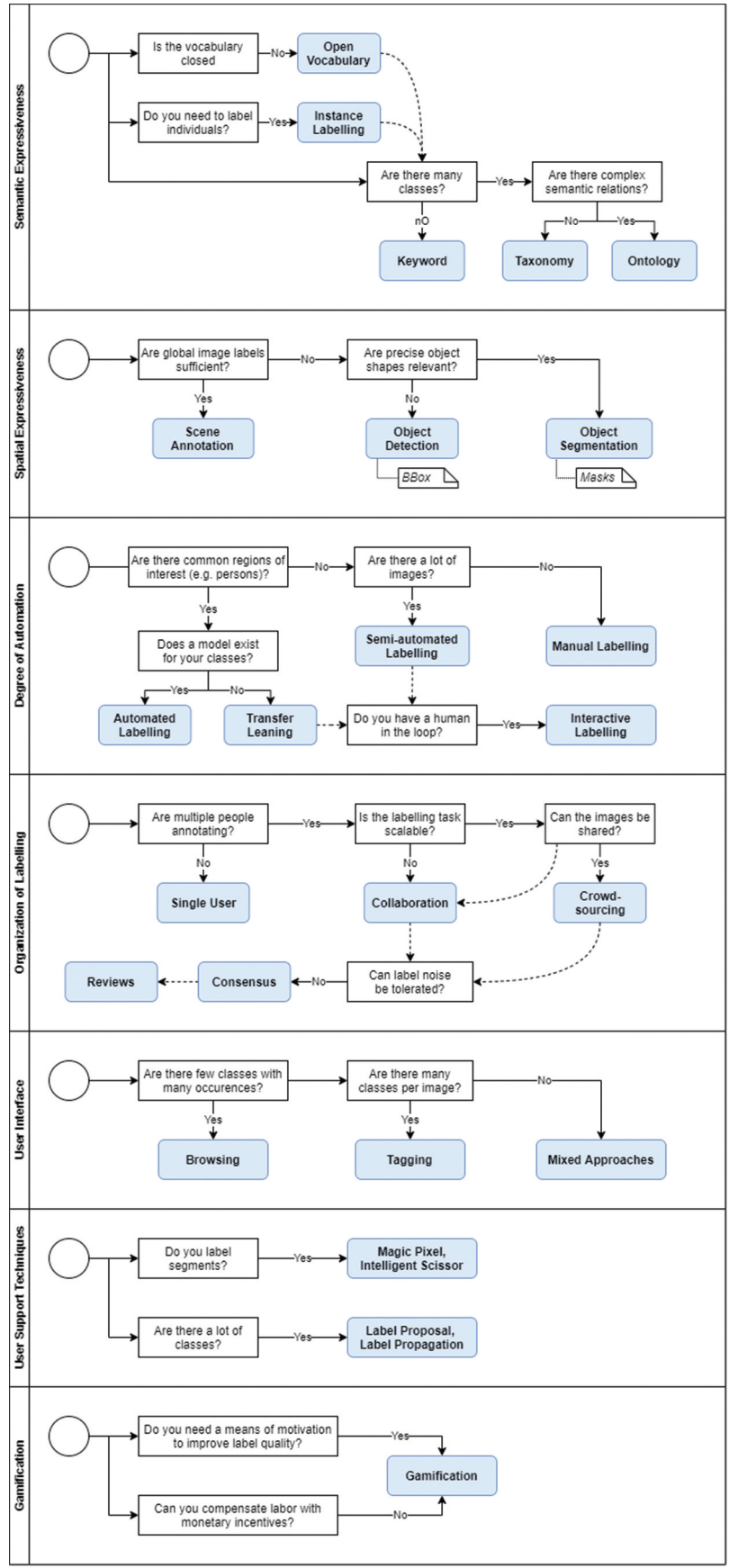

**Figure 9.** Decision tree for the selection of image labelling software.

### 5.2.5. Statistics generation

In some contexts, the automatic image analysis might be used to produce aggregated statistics. In this case, rare misclassifications are not of concern, but manual labelling effort is. Therefore, an initial model should be trained using semi-automatic or interactive labelling. Spatial annotations can further be used to measure object sizes or create heatmaps of common object positions. To accelerate the labelling process, the user can take advantage of magic pixel features and label proposals. An additional task is the visualisation of the raw annotation data into meaningful metrics, graphs, and statements.

### 5.3. Application domains

We found that some domains or industries have developed a need for specific applications of supervised ML that require the careful labelling of data. To illustrate the context for which image labelling is relevant, we highlight the most prevalent domains found in the following.

### 5.3.1. Autonomous driving

The labelling of frames (i.e., video sequences) is particularly labour-intensive but necessary to identify movement over time as it is required for autonomous driving (Grigorescu et al., 2020). For example, the Trajectory Annotation Tool allows refining an initial segmentation over all frames of a given sequence (Lladó et al., 2016). As the research in autonomous driving accelerates, there will be a need for efficient motion labelling. Schöning et al. (2015) provide an interactive solution for the interpolation of polygons over a video sequence. The user sets a seed annotation and is then being asked to correct if the semi-automatic annotations appear to be incorrect. Wang et al. (2018) describe a similar method and implement an object tracker with interactive human intervention. The semi-automatic segmentation of street scenes by Petrovai et al. (2017) generates even more detailed information to create annotated segments and reduce the annotation time by a factor of three compared to plain polygon drawing.

### 5.3.2. Healthcare

The use of automatic or semi-automatic diagnosis is especially common in the field of radiology and constitutes the sub-field CADx concerned with assisting physicians with image interpretation. For example, practitioners use software to evaluate X-ray images or computed tomography data to observe skin lesions or detect melanoma (Ferreira et al., 2012). Balducci and Borghi (2017) developed ILS for annotating skin images focusing on usability by adapting the interface to the end-user (physicians) and supporting touch pens. Mata et al. (2017) facilitate better consensus between experts for the detection of prostate cancer by analysing magnetic resonance images. Morrison et al. (2018) developed an ILS with semi-automated and fully automated labelling that reduces the evaluation time of cerebral microbleeds by a factor of five without reducing the accuracy. Son et al. (2018) see their objective not merely in automation but more in assisting practitioners. Their user interface provides a stepwise process for the assessment of fundus images. Mata et al. (2012) focus on reference and training. They developed an ILS for annotating mammography images with polygonal marks and notes supported by calculated metrics such as the area of the annotation, as well as features to look up similar cases in a database.

### 5.3.3. Mapping & surveillance

Disaster response requires immediate and targeted help. Salisbury et al. (2016) develop an ILS for real-time labelling of live aerial images by the crowd, aiming to rapidly identify points of interest and thus reducing the cognitive load of the pilots. Similarly, Zhuo et al. (2019) propose an algorithm to simplify the annotation of airborne images. They use existing map data to transfer labels from a given ground truth to their own dataset. The output of surveillance cameras comprises a large amount of video data, which only changes if objects walk through the scene. This poses the challenge to only identify relevant sequences. Kim et al. (2013) address this by developing a semi-automated video labelling software, which generates an initial annotation so that the user only has to check the validity. Zeinstra et al. (2017) focus on the context of forensics and person identification. They provide a dataset and various tools for analysing anthropomorphic features using landmarks, splines, and point clouds. Another application is the detection of specific behaviour patterns from surveillance cameras. Dufour et al. (2018) develop an ILS and image dataset to detect hazardous behaviour on ski lifts. They also consider the privacy aspect by anonymising people using face detection and tracking.

### 5.3.4. Research & sociology

Teams participating in robot competitions need a large pool of ground truth images to train robust cognition abilities for robots. Fiedler et al. (2019) developed their labelling specifically for RoboCup competitions. They focus on collaborative work but also introduce the possibility to include pre-trained models for initial annotation. This is particularly interesting if the competition scenario itself does not allow heavy models due to high inference time. There are already well-developed ontologies for biological species and plants, but a lack of fine-grained spatial annotations. Lingutla et al. (2014) present ILS for labelling image segments with ontologies and focus mainly on biological data. Social sciences widely use gestures in speeches and presentations to analyse behaviour. The manual annotation of hands is time-consuming and costly. Beugher et al. (2018) developed semi-automated ILS focusing on minimising the human workload while keeping a high accuracy. They use a confidence threshold as a metric for deciding if human intervention is needed. Similar tools are FreiHAND (Zimmermann et al., 2019) that supports the semi-automated labelling of hand poses and shapes, and HOnnotate (Hampali et al., 2020), which makes use of depth images.

### 5.3.5. Television

Aljundi et al. (2017) present a fully automated classification software for identifying actors in movies and series. An important aspect of video and especially television annotation is to keep track of objects or instances over time. Bianco et al. (2015) achieve this goal by adding a timeline for each object underneath the video display. The timeline shows the appearance of the object or person and if it was manually or automatically labelled. Another concept is instance exploring, where the user selects one specific instance and then is presented with all frames containing that instance to help the reviewer check that an actor is labelled correctly in all scenes. Another application is the automatic generation of meta data.

## 6. Conclusion and outlook

The technological achievements in the fields of AI and ML have strongly simplified the development of intelligent CV systems. Still, the big challenge remains that these systems need to be trained with large samples of annotated examples in a supervised manner. Here, the pivotal role of ILS comes into play to feed the underlying knowledge base with human-verified, high-quality data. In this sense, ILS constitute a central interface between intelligent machines and human professionals to project domain-specific knowledge into the learning base of CV systems. It can therefore be stated that the development and refinement of future CV systems will not only involve programmers and data science experts, but in particular business users and domain experts with their specific background (e.g., insurance workers to distinguish fraudulent from non-fraudulent cases or inventory managers to specify different types and qualities of incoming goods based on image and video material). Hence, it is important that not only developers and researchers are aware of the possibilities and functionalities of ILS, but also the group of potential business users, which will presumably grow even further in the near future.

To this end, our article systematises the field of image labelling and available ILS for CV applications to provide an understanding of the underlying concepts and features of image labelling and its implementation in software. We detailed annotation expressiveness, degree of automation, organisation of manual labelling, types of user interfaces, image labelling support techniques, as well as gamification. We provide a real-world benefit for research and business by providing a systematisation and overview of current ILS as well as guidance for ILS selection. Likewise, system vendors and developers can use the overview to add new functionalities to their tools and applications or be inspired for further improvements. Furthermore, using this schema, we discuss five dominant use archetypes (complex scene understanding, diagnosis and classification, image retrieval, instance identification, and statistics generation) as well as present the key application areas of autonomous driving, healthcare, mapping & surveillance, research, and television.

As a limitation, our current study is of qualitative and summative nature. Thus, in the next steps, our derived schema could be enhanced as a schema for ILS assessment and quantitative decision making by assigning decision weights to the identified dimensions and characteristics. Likewise, future research should consider a systematic quantitative evaluation based on standardised benchmark datasets to allow for an ILS comparison based on labelling performance. Reasonable evaluation metrics could be the labelling time per image in seconds, the intersection over union to measure the label precision, and the user satisfaction.

Besides, the emergence of interactive approaches shows great potential for more efficient labelling resulting in various future research directions. Especially in healthcare, the joint decision making and labelling between humans and machines (also called hybrid intelligence, Zschech et al., 2021) yields significant potential. Further, methods of active or reinforcement learning might, in the long run, reduce the amount of manual labelling. However, new media types, such as depth images or point cloud data, could lead to new challenges in labelling tasks (e.g., Sager et al., 2021). Economic considerations of labour costs (crowdsourcing) and opportunity costs (misclassification) could allow the calculation of an optimal cost-accuracy trade-off.

In consequence, an industry of service providers has emerged already and will further proliferate, ranging from simple labelling platforms through outsourcing services to so-called "full-stack AI"-platforms offering the benefits of transfer learning (Janiesch et al., 2021), comprising a vast and diverse landscape for further research.

## Notes

1. http://www.image-net.org/
2. The core of this survey is on software for the labelling of individual photos and other images such as photos,

scans, or frames. Methods and tools for analysing 3D objects or moving images are not in focus for this survey, although we may touch upon these topics at times.
3. For video-based image labelling techniques see Appendix A.3.

## Disclosure statement

No potential conflict of interest was reported by the authors.

## Funding

This work was supported by the Bayerische Staatsministerium für Wirtschaft, Landesentwicklung und Energie (StMWi) [DIK0143/02].

## ORCID

Christoph Sager http://orcid.org/0000-0002-7060-0541
Christian Janiesch http://orcid.org/0000-0002-8050-123X
Patrick Zschech http://orcid.org/0000-0002-1105-8086